\documentclass[a4paper,conference]{IEEEtran}

\usepackage{graphicx,subfig}
\usepackage{comment}
\usepackage[font=small]{caption}
\usepackage{multicol}
\usepackage{cite}

\usepackage[cmex10]{amsmath}
\usepackage{amssymb}
\usepackage{algorithmic}
\usepackage{algorithm}
\usepackage{xcolor,mathrsfs,bbm}

\usepackage{url,float}
\usepackage{cases}
\usepackage{enumitem}

\newcommand{\ie}{\textit{i.e.}}
\newcommand{\et}{\textit{et al}}
\newcommand{\p}{^{\prime}}

\ifCLASSINFOpdf
\else
\fi

\begin{document}
%
\title{Weakly Supervised Geodesic Segmentation of Egyptian Mummy CT Scans}



%
\author{\IEEEauthorblockN{Avik Hati\IEEEauthorrefmark{1},
Matteo Bustreo\IEEEauthorrefmark{1},
Diego Sona\IEEEauthorrefmark{1}\IEEEauthorrefmark{4},
Vittorio Murino\IEEEauthorrefmark{1}\IEEEauthorrefmark{2}\IEEEauthorrefmark{3} and
 Alessio {Del Bue}\IEEEauthorrefmark{1}\\
Avik.Hati17@gmail.com, \{Matteo.Bustreo, Diego.Sona, Vittorio.Murino, Alessio.DelBue\}@iit.it}
\IEEEauthorblockA{\IEEEauthorrefmark{1}Pattern Analysis and Computer Vision, Istituto Italiano di Tecnologia, Genova, Italy}
\IEEEauthorblockA{\IEEEauthorrefmark{4}Neuroinformatics Laboratory, Fondazione Bruno Kessler, Trento, Italy}
\IEEEauthorblockA{\IEEEauthorrefmark{3}Huawei Technologies Ltd., Ireland Research Center, Dublin, Ireland}
\IEEEauthorblockA{\IEEEauthorrefmark{2}Department of Computer Science, Universit\`a di Verona, Italy}}

\maketitle

\begin{abstract}
In this paper, we tackle the task of automatically analyzing 3D volumetric scans obtained from computed tomography (CT) devices. In particular, we address a particular task for which data is very limited: the segmentation of ancient Egyptian mummies CT scans. We aim at digitally unwrapping the mummy and identify different segments such as body, bandages and  jewelry. The problem is complex because of the lack of annotated data for the different semantic regions to segment, thus discouraging the use of strongly supervised approaches. We, therefore, propose a weakly supervised and efficient interactive segmentation method to solve this challenging problem. After segmenting the wrapped mummy from its exterior region using histogram analysis and template matching, we first design a voxel distance measure to find an approximate solution for the body and bandage segments. Here, we use geodesic distances since voxel features as well as spatial relationship among voxels is incorporated in this measure. Next, we refine the solution using a GrabCut based segmentation together with a tracking method on the slices of the scan that assigns labels to different regions in the volume, using limited supervision in the form of scribbles drawn by the user. The efficiency of the proposed method is demonstrated using visualizations and validated through quantitative measures and qualitative unwrapping of the mummy.
\end{abstract}


%
\IEEEpeerreviewmaketitle

\section{Introduction}
\label{sec:intro}
\begin{figure}
    \centering
    \includegraphics[width=0.99\linewidth, clip,trim={0 10mm 0 9mm}]{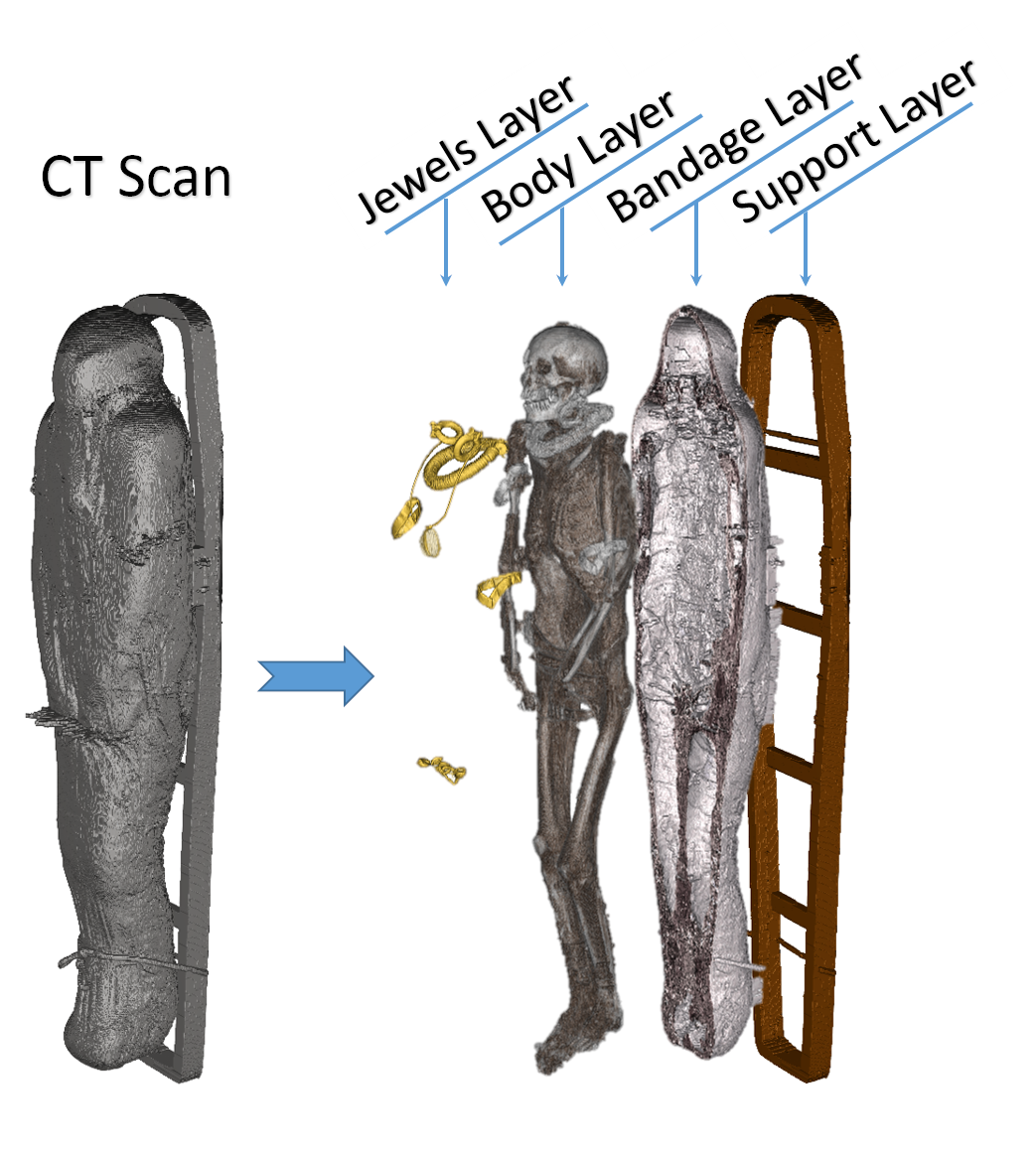}
    \caption{Mummy CT scan \cite{mummy_ME_collection} and its segmentation in \emph{Exterior objects of support}, \emph{Bandage}, \emph{Body} and \emph{Metal} obtained by applying the proposed method.}
    \label{fig:problem}
\end{figure}

Analysis of ancient mummies and their funerary equipment is an important task in archaeological studies as they represent an extremely well preserved and authentic view of the past. Since half of the 19th century, mummies were scientifically investigated by unwrapping the tape or bandages, and removing amulets and jewels from the body \cite{riggs2014unwrapping}. However, this process is devastating and irreversible, therefore it is desirable to avoid it whenever possible.
%
%
Hence, it is essential to build a mechanism that can acquire information about what is hidden by the bandages without unwrapping the mummy and provide this information to the archaeological scholars. 3D volumetric scans obtained from CT devices are the most promising data for these cultural heritage studies~\cite{casali2006x, hughes2011ct, cesarani2009multidetector, curto1968news}.
In this work, we propose a novel segmentation algorithm and apply to 3D CT scan of the entire mummy for digitally unwrapping its bandage and visualize its body and other objects present (Fig.~\ref{fig:problem}).
%
%
%

Even though our task is unique, CT scans are standard data used in medical image processing \cite{wang2018deepigeos}. Several unsupervised, fully-supervised and semi-supervised methods have been proposed in the literature.
However, both unsupervised \cite{sharma2010automated} and fully-supervised \cite{shelhamer2017fully} methods are not suitable for the problem of 3D mummy CT scans segmentation. In fact, mummy's tissues do not have the same characteristics of typical medical data. Hence, labeled data are not available for training, and it is extremely difficult to get manual annotations since it is typically a laborious and time-consuming activity requiring an expert.
Further, mummy’s body region is not the most salient \cite{zhu2014saliency} part of the CT data, making its segmentation very difficult using unsupervised techniques.
Because of these reasons, to solve the mummy segmentation problem, we propose a weakly supervised method which requires less user interaction than fully supervised methods, while achieving higher accuracy and robustness compared to unsupervised methods \cite{zhao2013overview}. 

\begin{figure}
    \centering
    \includegraphics[width=0.65\linewidth]{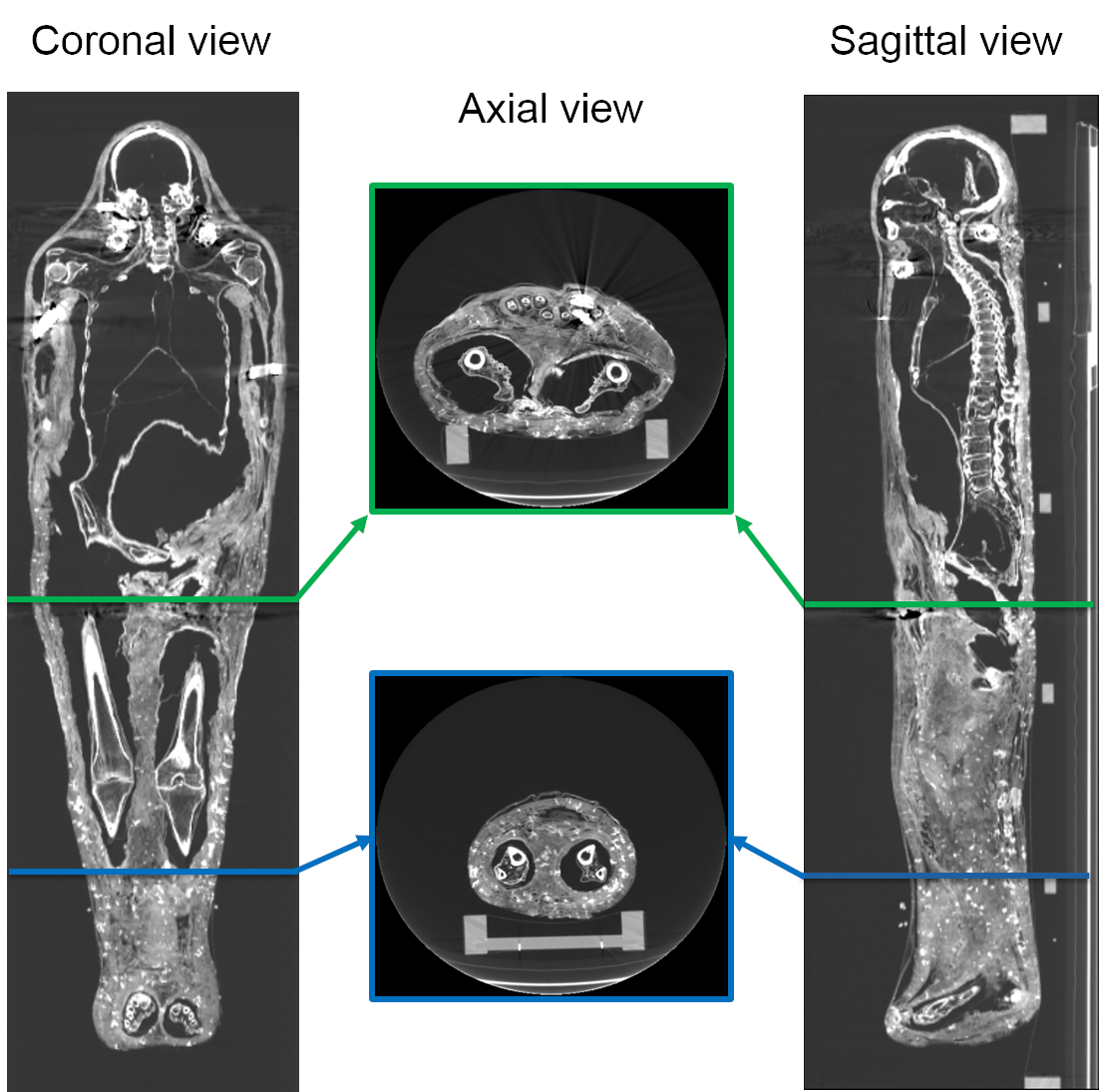}
    \caption{Example of coronal, axial and sagittal views of the CT scan of a human mummy \cite{mummy_ME_collection} (contrast enhanced for better visualization).}
    \label{fig:kha_views}
\end{figure}

In this paper, 
given a CT scan of a mummy, we  segment the data into multiple volumetric semantic parts including bandage, body, metal (\textit{e.g.} jewelry) and other exterior objects, such as the one depicted in Fig~\ref{fig:problem}. 
To accomplish this task, we propose a two-stage weakly supervised algorithm:
\begin{enumerate}
\item In the first stage, after detecting the regions exterior to the mummy using prior knowledge about its contents \textit{e.g.} air and wooden support, we use geodesic distance measure to obtain an approximate segmentation of the mummy into body and bandages, in an unsupervised framework.
\item In the second stage, we apply GrabCut and segment tracking methods to obtain the final segmentation. During this stage, we make use of human interaction in the form of scribbles for refining the result.
\end{enumerate}
In Fig.~\ref{fig:kha_views}, we show a qualitative example of our input data, \textit{i.e.} the scan of an ancient human mummy from coronal, sagittal and axial viewpoints, and a detailed description is reported in Sec.~\ref{ssec:data}. Here, we would like to stress upon the complexity of the data. It is evident from Fig.~\ref{fig:kha_views} that at some locations, bandage tissue is adjacent to the body while both having very similar radiodensity, which makes the problem challenging. 

To summarize our contribution, we propose an efficient interactive segmentation method, with limited user interaction needed: user intervention is required only at the final stage of the algorithm, for result refinement. Even though applied to this specific case, since supervision is minimal, it can be applied to other cases for which annotated data is not available.
To the best of our knowledge, this is the first work attempting to solve the problem of mummy 3D CT scan segmentation by a weakly supervised method.

\section{Related works}
\label{sec:lit}
In the next sections, we review relevant works on application of CT scans to cultural heritage studies and CT scan analysis in medical image segmentation.

\subsection{CT scan analysis in cultural heritage}
In the cultural heritage field, CT scans are frequently used to analyze archaeological samples.
Albertin \et. 
\cite{albertin2016virtual} proposed virtual reading and digitization of text in ancient books and handwritten documents from their 3D tomographic images. Presence of metals (\textit{e.g.} iron, calcium) in ancient inks \cite{stromer2018browsing} produces sufficient contrast between inked text and the paper in the images.
Similar virtual unwrapping of fragile scroll fragments has been explored in \cite{seales2016damage}.
Another collection of works emphasize on obtaining details of archaeological objects' (\textit{e.g.} sculptures) inner structures \cite{brancaccio2011real,re2014x}. This helps in proper planning for their conservation and restoration as well as understanding historical artefact creation techniques. Typically a CT volume of the object is reconstructed and visualized for its analysis 
\cite{huppertz2009nondestructive}. 
Works in \cite{martina2005kha, bianucci2015shedding} used CT scans of mummies to study their general conditions, pathology, embalming and bandaging techniques as well as to identify the bandaged species in case of animals.
However, the scope of these works are limited to manually studying the 3D visualization. In the given task, we  propose a semi-automatic method to segment different regions in the mummy CT data.
\begin{figure}[b]
    \centering
    \subfloat[]{
    \includegraphics[height = 0.3\linewidth, clip, trim={0 0 0 0mm}]{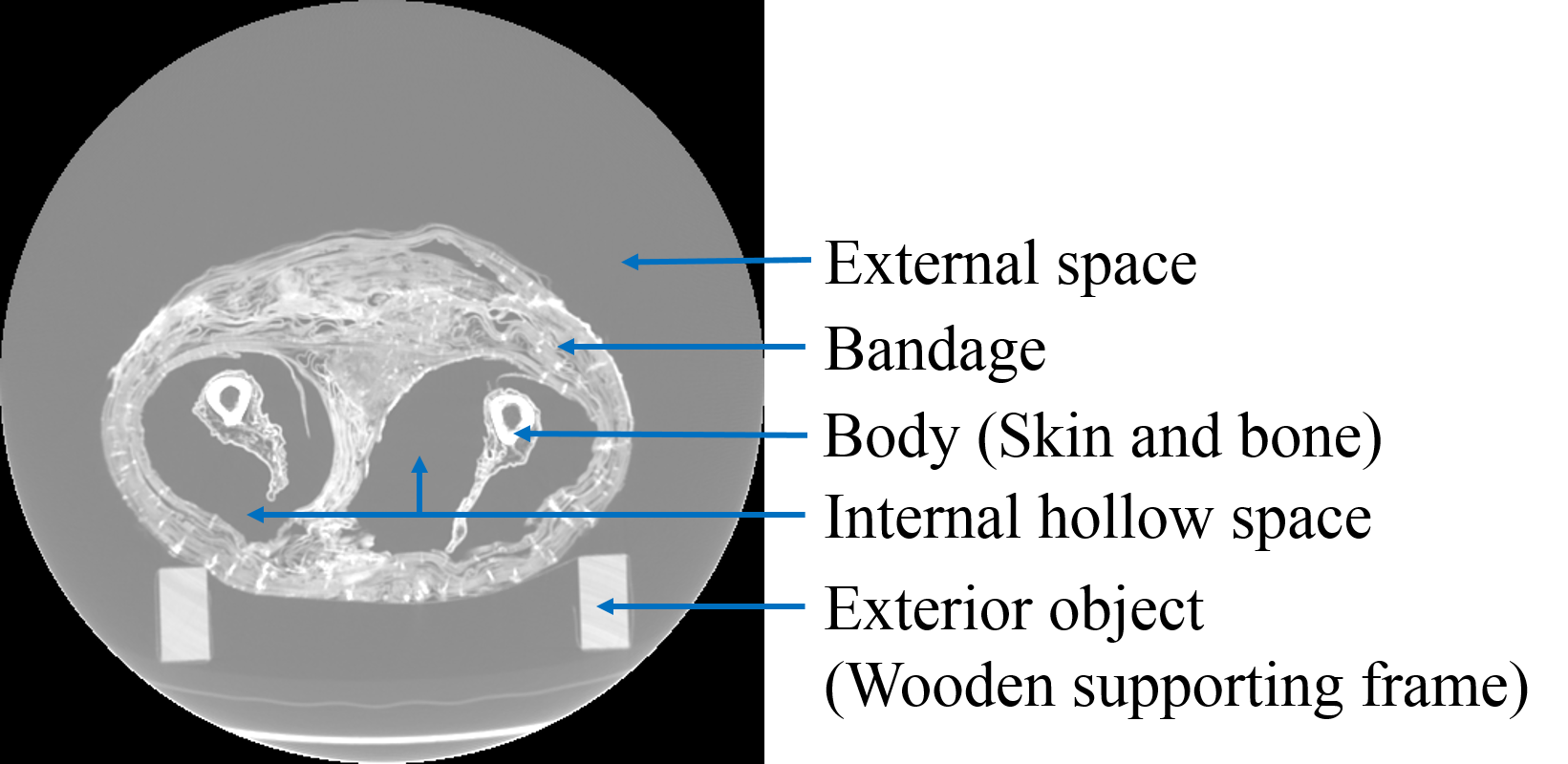}
    \label{fig:data}}
    \subfloat[]{
    \includegraphics[height = 0.3\linewidth, clip, trim={0 0 0 0mm}]{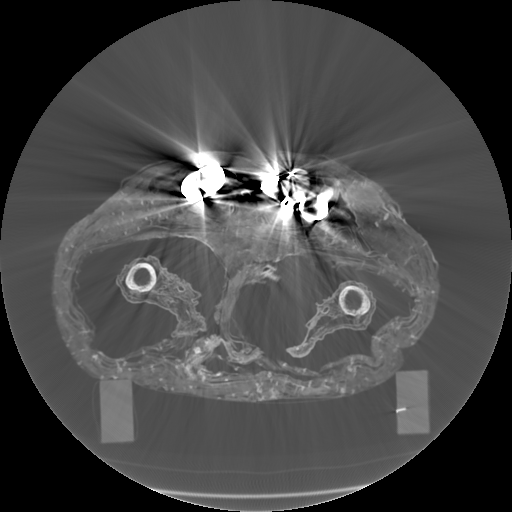}
    \label{fig:artifact}}
    \caption{Visualization of axial frames of a mummy \cite{mummy_ME_collection} selected in proximity of the thigh. (a) Some of the regions we are interested to segment are indicated. (b) Artifacts caused by presence of metals.}
\end{figure}
%

\subsection{CT scan analysis in medical image segmentation}
In the medical image segmentation field, one of the most important problems is the segmentation of anatomical structures for medical diagnosis \cite{wang2018deepigeos}. Noise in the data and variations in data acquisition platforms make extremely challenging for unsupervised methods to perform well~\cite{sharma2010automated}. Recent evolution in supervised deep convolutional neural networks (CNN) \cite{shelhamer2017fully} overcame these limitations and obtain high quality segmentation results. CNN-based methods such as U-Net \cite{ronneberger2015u}, V-Net \cite{milletari2016v} 
and DeepMedic \cite{kamnitsas2017efficient} have been shown to attain state-of-the-art performance. However, in case of scarcity in training data, training of such networks is hardly feasible. For this reason, semi-supervised methods are more useful.

These interactive methods can incorporate task-specific knowledge, thus
performing better than automated approaches \cite{borovec2017supervised}. Interactions can be provided in the way of clicks \cite{haider2015single}, contours \cite{xu1998snakes}, bounding boxes~\cite{rother2004grabcut} or scribbles \cite{bai2007geodesic}.
However, most of these methods require a large amount of user interaction and they frequently focus on segmenting only specific parts of the body. For example, Rajchl \et. \cite{rajchl2016deepcut} used bounding box annotation and solved a conditional random field based energy minimization problem to segment brain and lungs. Level set based methods in \cite{yushkevich2006user,cates2004gist} use manual labeling of regions of interest 
for segmenting different regions in the brain. These methods apply annotations at the beginning of their respective pipelines and do not have the flexibility to refine the results at later stages. Wang \et. \cite{wang2016slic} proposed a method to segment placenta and it learns features of the entire organ by propagating foreground-background scribbles drawn on one or few slices. However, for mummy segmentation, we need to process the entire mummy which is very challenging. Further, our method automatically finds the annotation required in the initial stage and requires very few user annotation during the refinement stage.
%

\section{The proposed method}
\label{sec:method}
\begin{figure}[b]
    \centering
    \includegraphics[width = 0.9\linewidth]{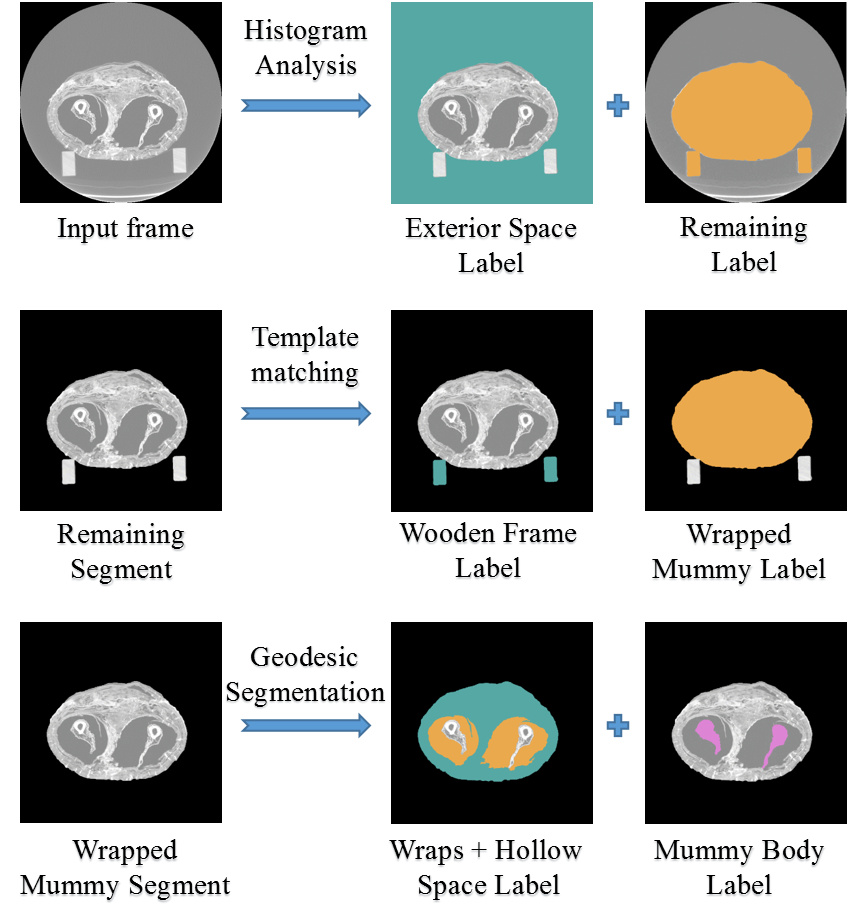}
    \caption{Hierarchical segmentation of the mummy's CT scan.}
    \label{fig:flow}
\end{figure}
%

\subsection{Data representation}
\label{ssec:data}
The mummy CT scan input data is a volume represented as a sequence of 2D images, where each image is a fixed thickness axial slice (frame) of the mummy. 
The value of each image voxel represents the measured radiodensity, reported using the Hounsfield Unit (HU) scale. 
Our goal is to group and segment each voxel into the following regions:
\begin{itemize}
\item \emph{Bandage} which is used to wrap the body;
\item \emph{Body} of the mummy, comprised of skin and bone;
\item \emph{Metals} (typically jewelries), which lie over the body;
\item \emph{Interior hollow space}, which is the gap between the wrapping bandages and the body. This space has been created due to the shrinking of the body over the years;
\item \emph{External space}, which is the vacuum around the mummy acquired during scanning;
\item \emph{Exterior objects}, such as structures supporting the mummy (\textit{e.g.} wooden support).
\end{itemize}
In Fig.~\ref{fig:data}, we show one axial slice of a male mummy with indication of the regions to be segmented.
Fig.~\ref{fig:artifact} illustrates that metals present inside a bandaged mummy generate severe artifacts, which are not present in standard biomedical data.
%

\subsection{Overview of the approach}
\label{ssec:summary}
For segmenting the input CT scan into multiple volumetric semantic parts, we adopt a hierarchical approach and we segment the regions in a sequential manner, as illustrated in Fig.~\ref{fig:flow}. Given a sequence of 2D images (axial frames), we pre-process every image independently for identifying all the frame voxels which are not related to the mummy, \textit{i.e.}  \emph{External space}, \emph{Exterior objects} and \emph{Metals}. We achieve this using a combination of histogram equalization, template matching, Hough transform and connected component analysis, as described in detail in Sec.~\ref{ssec:stage0}. 
The identified set of voxels is then used as reference in the computation of geodesic distances for obtaining an approximate segmentation of the remaining voxels in \emph{Body} and \emph{Bandage}, as detailed in Sec.~\ref{ssec:stage1}. 
The final stage of the algorithm consists of GrabCut based segmentation and inter-frame segments tracking for updating and improving the results obtained with the previous stage. Unlike the previous stages where each slice is processed independently, in this stage we process the full volume jointly, as described in detail in Sec.~\ref{ssec:stage2}.
In the following sections, we provide additional details about the implemented method.

\subsection{Pre-processing}
\label{ssec:stage0}
In pre-processing stage we aim at segmenting all the elements which are not related to the wrapped mummy's body.

Separating the \emph{External space} (first row in Fig.~\ref{fig:flow}) in most of the cases is a straightforward operation, since it is typically composed of air, and therefore it has a known Hounsfield Unit value. In order to ensure a proper segmentation also in the cases in which CT scan values have been scaled or shifted, we use histogram analysis for choosing the threshold better separating voxels with low radiodensity (air) from voxels with higher radiodensity (wrapped mummy or wooden support).

Most of the times, mummies lie on a support (\emph{Exterior object}) which needs to be separated from the body. Mummy supports have regular and known shapes in the CT scans we have. Because of this, we apply template matching for detecting them, and optimize the process by locating vertical edges using Hough transform (second row in Fig.~\ref{fig:flow}). 

Further, it is also known that \emph{Metals} have Hounsfield Unit value larger than any other expected element in the scene. Thus, we find the jewelries present in the mummy by thresholding the voxels and grouping connected components. 

The visualizations of the components segmented in this stage are reported in Sec.~\ref{ssec:quali} (Fig. \ref{fig:visualize_wrap}, Fig. \ref{fig:visualize_gold}).

\subsection{Geodesic segmentation}
\label{ssec:stage1}
\begin{figure}[b]
    \centering
    \includegraphics[width = 0.40\linewidth]{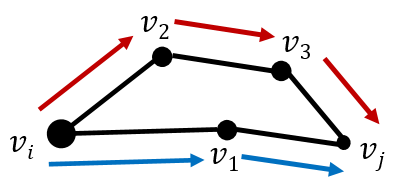}
    \caption{Illustration of geodesic distance computation in a graph between nodes $v_i$ and $v_j$. It is defined as the minimum cumulative distance along two paths indicated in red and blue. $d_g(v_i,v_j) = \min \{ d(v_i,v_1)+d(v_1,v_j), d(v_i,v_2)+d(v_2,v_3)+d(v_3,v_j) \}$}
    \label{fig:geodesic}
\end{figure}

The mummy's \emph{Body} consists of several components (mainly bones and skin but internal organs can also be observed). Wrapping bandages are non-uniform and their density is not constant, possibly also because of degradation due to ageing. Hence, the regions we want to segment are non-homogeneous and they are frequently disconnected.
Moreover, \emph{Bandage} and \emph{Body} have, overall, a very similar radiodensity and it is not uncommon that two adjacent and similar voxels belong to the two different segments. Thus, using the standard Euclidean distance for separating voxels belonging to different regions is not effective. We can, nevertheless, leverage the data structure of the mummy's CT scan: \textit{Body} is wrapped in \textit{Bandage} which is surrounded by the exterior region (\emph{External space} and \emph{Exterior object}), as illustrated in Fig.~\ref{fig:data}. Hence, we decided to calculate geodesic distance from the exterior region to the unclassified voxels for separating the ones belonging to the \textit{Body} and the ones belonging to the \textit{Bandage}.\\ 
%
%
%
\textbf{Geodesic distance.}
Consider an undirected graph $\mathcal{G} = (\mathcal{V}, \mathcal{E})$ with set of nodes $\mathcal{V} = \{v_i\}$ and set of edges $\mathcal{E} = \{e_{ij}\}$. Let $\mathrm{P}^{ij} = \{\mathcal{P}_1^{ij}, \mathcal{P}_2^{ij}, \mathcal{P}_3^{ij}, \ldots \}$ denote the set of all paths from node $v_i$ to node $v_j$ that can be traversed in the graph. A path $\mathcal{P}^{ij}$ is a set of edges corresponding to a sequence of node pairs $(v_0, v_1), (v_1, v_2), \ldots, (v_{n-1}, v_n)$ where $v_0 = v_i$, $v_n = v_j$ and $n$ is the length of the path. Given any distance measure $d(\cdot)$ (\textit{e.g.} Euclidean distance), the geodesic distance $d_g(\cdot)$ between a pair of nodes $v_i$ and $v_j$ is defined as the cumulative pairwise distance along the shortest path as follows: 
\begin{equation}
\label{eqn:geodesic}
d_g(v_i, v_j) = \underset{\mathcal{P}^{ij} \in \mathrm{P}^{ij}}\min \sum_{(v_k, v_{k+1}) \in \mathcal{P}^{ij}}{d(v_k, v_{k+1})} \,,
\end{equation}
and it is illustrated in Fig.~\ref{fig:geodesic} using an example graph.

In the proposed method, we divided each axial frame into $3\times3$ patches representing the nodes of the graph and we used the patch average voxel radiodensity as feature for computing the distance $d(v_k, v_{k+1})$ in Eq.~(\ref{eqn:geodesic}).
%
%
Note that, instead of patches, it is also possible to use superpixels or regions obtained by any over-segmentation method. 

Let us define the set of voxels belonging to the \emph{Body} as $\mathrm{R}_b$, those belonging to the wrapping \emph{Bandages} as $\mathrm{R}_w$ and those belonging to either the \emph{External space} or the \emph{Exterior objects}, detected in Sec.~\ref{ssec:stage0}, as $\mathrm{R}_e$. At this stage of the algorithm, we consider the voxels belonging to the \emph{Bandages} and the \emph{Body} together
to be part of the same set $\mathrm{R}_{w+b}$. 
In order to segment $\mathrm{R}_{w+b}$, we first compute average geodesic distance of every node $v_i \in \mathrm{R}_{w+b}$ from reference region $\mathrm{R}_e$ as follows:
%
%
\begin{equation}
\label{eqn:avg_dist}
\bar d_g(v_i, \mathrm{R}_e) = \frac{1}{|\mathrm{R}_e|} \sum_{v_j \in \mathrm{R}_e}{d_g(v_i, v_j)} \,.
\end{equation}

We observe that the reference region contains some noisy voxels with significantly high and low intensity values and they may have a negative influence in the averaging operation in Eq.~(\ref{eqn:avg_dist}). To achieve efficiency and better performance, we modify the formulation in Eq.~(\ref{eqn:avg_dist}) in order to consider only the smallest $m$ geodesic distances $d_g(\cdot)$ between node $v_i \in \mathrm{R}_{w+b}$ and the graph nodes $v_j$ in $\mathrm{R}_e$:
\begin{equation}
\label{eqn:avg_dist_top}
\bar d_g(v_i, \mathrm{R}_e) = \frac{1}{m} \sum_{v_j \in \mathrm{R}_e}{d_g(v_i, v_j) \mathrm{I}(d_g(v_i, v_j) \leq \bar D^i)},
\end{equation}
where $\mathrm{I}(\cdot)$ is an indicator function and $\bar D^i$ is the $m$-th minimum geodesic distance computed for node $v_i$. For notation simplicity, let us denote $\bar d_g(v_i, \mathrm{R}_e)$ as $d_i$.
%
%
%
\begin{figure}
\belowcaptionskip1ex
\begin{multicols}{3}
\begin{tabular}{@{}c}
    \subfloat[]{\includegraphics[width = 0.9\linewidth]{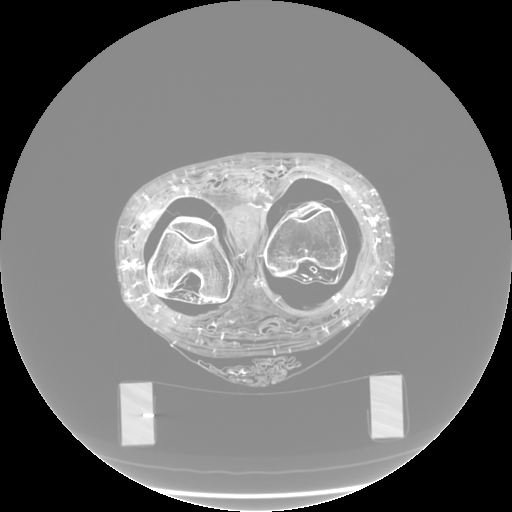}}
    \\
    \subfloat[]{\includegraphics[width = 0.9\linewidth]{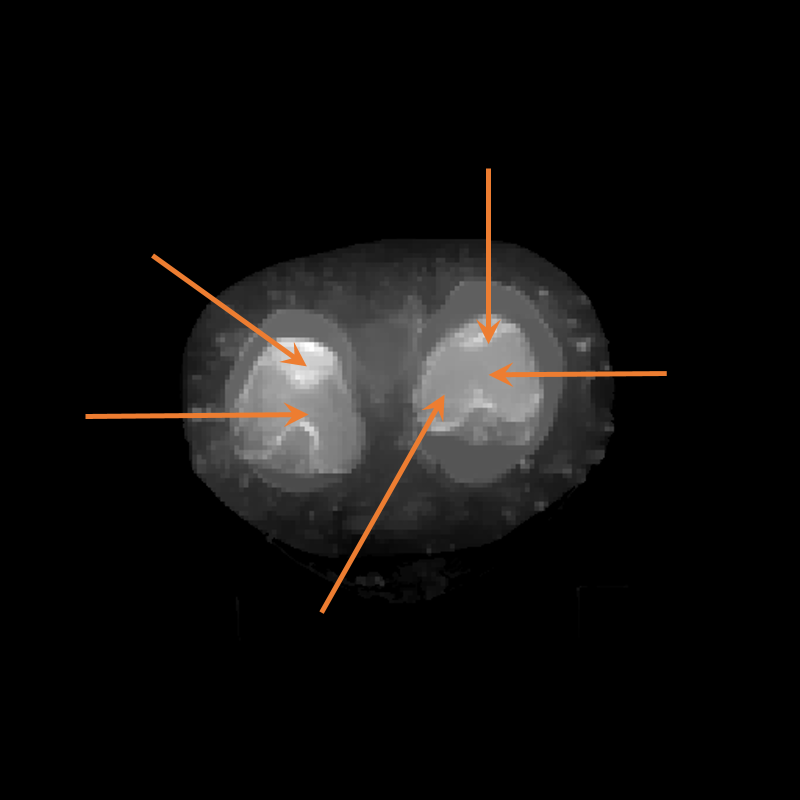}}
    \label{fig:sfig2}
\end{tabular}

\subfloat[]{\includegraphics[width=2\linewidth]{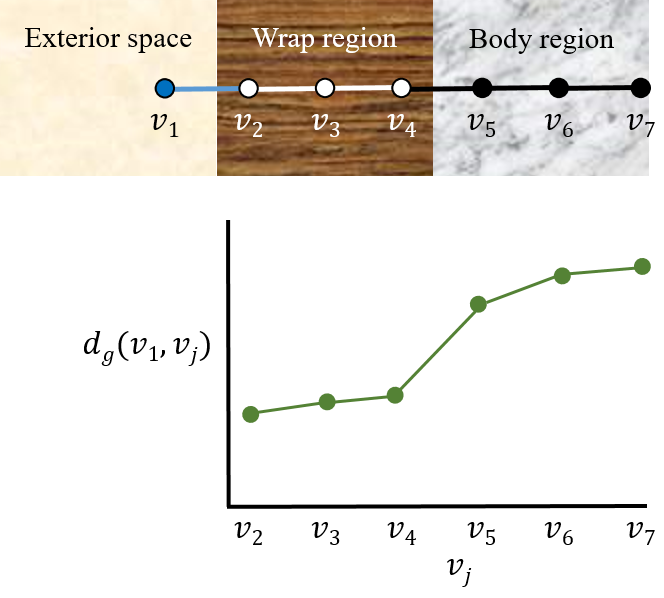}\vspace*{\fill}}
\end{multicols}
\caption{Illustration of geodesic distance. (a) Input image. (b) Average geodesic distance $\bar d_g$ (Eq.~(\ref{eqn:avg_dist_top})) of every voxel from the external space $\mathrm{R}_e$. Arrows indicate that $\bar d_g$ increases with traversal from the wrapping bandage region to the body region. (c) Illustration using a synthetic image. $d_g(v_1,v_5)$ is significantly larger than $d_g(v_1,v_4)$ whereas $d_g(v_1,v_4)$ is slightly larger than $d_g(v_1,v_2)$ and $d_g(v_1,v_3)$ because $d(v_4,v_5)$ is larger than remaining pairwise distance $d(\cdot)$.}
\label{fig:geodesic_map}
\end{figure}
%
%
%

We argue that patches with smaller geodesic distances constitute the mummy wrap region ($\mathrm{R}_w$) and patches with larger geodesic distances constitute the body region ($\mathrm{R}_b$). This is because computation of geodesic distance between a node pair $( v_i \in \mathrm{R}_b, v_j \in \mathrm{R}_e )$ requires a larger path to be traversed than computing the same for $( v_i \in \mathrm{R}_w, v_j \in \mathrm{R}_e )$. 
Hence, without loss of generality, we expect a significant change in gradient in the sequence of distances $d_i$ when arranged in ascending order, denoted as $\tilde d_i$. It is therefore possible to identify a value $m_2$ separating the (not necessarily connected) voxels belonging to $\mathrm{R}_w$ and $\mathrm{R}_b$:
\begin{equation}
\label{eqn:m2}
    m_2 = \arg \underset{i}{\max}\,{\frac{\partial \tilde d_i}{\partial i}} \,,
\end{equation}
\begin{equation}
\label{eqn:Rw}
    \mathrm{R}_w = \{ v_i \in \mathrm{R}_{w+b} : \tilde d_i \leq \tilde d_{m_2} \} \,,
\end{equation}
\begin{equation}
\label{eqn:Rb}
    \mathrm{R}_b = \{ v_i \in \mathrm{R}_{w+b} : \tilde d_i > \tilde d_{m_2} \} \,.
\end{equation}

In Fig.~\ref{fig:geodesic_map}, we illustrate the described process using a synthetic image and a real axial frame.
Note that in most slices, there exists an interior hollow space between the wrapping \emph{Bandage} and \emph{Body}. Detection of this region is trivial since its voxel values are the same as the air voxels in the detected exterior region.
In practice, the obtained set $\mathrm{R}_b$ may also include some patches from the wrap region in addition to the body patches. In the next section, we refine this result.

\subsection{GrabCut segmentation and tracking}
\label{ssec:stage2}
In Sec.~\ref{ssec:stage1}, we process every frame in the volume independently and obtain a coarse segmentation of different regions ($\mathrm{R}_e$, $\mathrm{R}_w$, $\mathrm{R}_b$). In this section, we update this result through volume level processing of data, ensuring information propagation across frames and thus yielding more accurate segmentation results.

\subsubsection{GrabCut}
\label{s2sec:grabcut}
Rother \et.~\cite{rother2004grabcut} introduced the GrabCut method for segmentation of objects of interest in an image. The algorithm separates the pixels into two categories: foreground and background. Typically, a part of the image is hard-labeled by user as background and the algorithm iteratively assigns labels to the remaining pixels. 
%
%
However, we can benefit from the fact that we don't need manual intervention in this stage and we use the output of the previous processing stage to feed the GrabCut algorithm. Specifically, since the goal of this stage is refining the region $\mathrm{R}_b$ obtained using Eq.~(\ref{eqn:Rb}), we initialize the background region with the set of voxels belonging to $\mathrm{R}_w \cup \mathrm{R}_e$.
The voxels labeled as foreground at the output of GrabCut constitute the updated \emph{Body} region $\mathrm{R}_b$.

Here, we apply GrabCut to volumes. However, segmenting the entire CT volume at one go is memory inefficient. Hence, we first split it into overlapping volumes, each constituted by $n_G$ successive frames, and segment them using GrabCut. The obtained results are then averaged to obtain the final segmentation mask. Using this approach, we ensure that the resulting mask changes smoothly across successive frames.
%
%
%

\subsubsection{Tracking}
\label{s2sec:tracking}
Since the data contains some voxels from different regions with similar appearances, ambiguity in the segmentation is relevant. 
GrabCut improves the result obtained in Sec.~\ref{ssec:stage1}, but we can further refine it by cross-slice matching of segments belonging to $\mathrm{R}_b$ across frames. We refer to this process as `tracking’ and the set of matched segments in a certain number of successive frames as a `track’. 
During the process, the segments which cannot be properly tracked are discarded from the final result of $\mathrm{R}_b$ and appended to $\mathrm{R}_w$, thus obtaining a more refined output than the GrabCut result.
\begin{figure}
    \centering
    \includegraphics[width=0.8\linewidth]{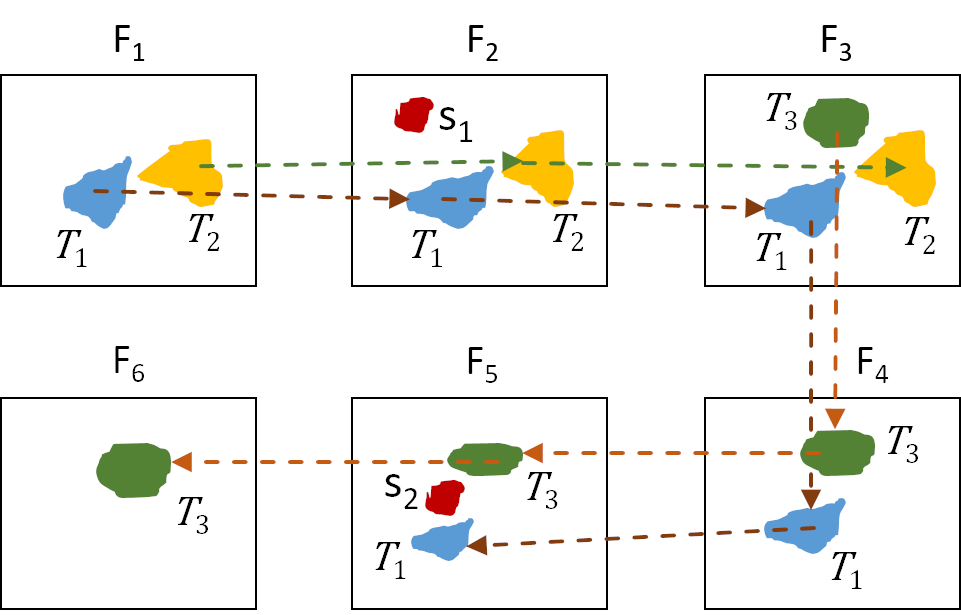}
    \caption{Illustration of tracking algorithm. User initializes tracks $T_1$ (\textit{blue} segment) and $T_2$ (\textit{yellow} segment) in frame-1 and track $T_3$ in frame-3 (\textit{green} segment), and they are tracked across subsequent frames. $T_1$ and $T_2$ cease in frame-5 and frame-3, respectively. Segments in all the frames are detected at the output of the GrabCut stage. Tracking algorithm refines that result by removing the spurious (\textit{red}) segments $\mathrm{s}_1$ in frame-2 and $\mathrm{s}_2$ in frame-5. }
    \label{fig:tracking}
\end{figure}

Let $L^{(k)}$ be the number of tracks in frame-$k$, $T_i^{(k)}$ denotes the segment corresponding to the $i$-th track in frame-$k$ and $H_i^{(k)}$ denotes an ensemble of segments in this track over past $M$ frames such as:
\begin{equation}
\label{eqn:track_history}
    H_i^{(k)} = [T_i^{(k)}, T_i^{(k-1)}, \ldots, T_i^{(k-M+1)}],\, \forall i=1,2,\ldots,L^{(k)} \,.
\end{equation}
In frame-$k+1$, a segment $s_j$ is added to the $i$-th track if it has the highest similarity with $H_i^{(k)}$. Thus $T_i^{(k)}$ grows as:
\begin{equation}
\label{eqn:track_update}
    T_i^{(k+1)} = \arg \underset{\mathrm{s}_j \in \mathrm{R}_b^{(k+1)}}{\max}\,\Phi({H_i^{(k)}, \mathrm{s}_j}) \,,
\end{equation}
where $\Phi(\cdot)$ is the similarity function that computes average similarities between the feature of segment $s_j$ and all segments belonging to a track over past $M$ frames, \ie, $H_i^{(k)}$. Here we consider spatial coordinates of all pixels in each segment as features. Further, the use of $H_i^{(k)}$ instead of $T_i^{(k)}$ while computing $\Phi(\cdot)$ using Eq.~(\ref{eqn:track_update}) ensures consistency in matching.

This process begins by a user identifying $L^{(1)}$ valid segments in frame-1 belonging to $\mathrm{R}_b$ and they initialize tracks $T_1^{(1)}, T_2^{(1)}, \ldots, T_{L^{(1)}}^{(1)}$, and define: 
\begin{equation}
\label{eqn:track_init}
    H_i^{(1)} = [T_i^{(1)}, T_i^{(1)}, \ldots, T_i^{(1)}],\, \forall i=1,2,\ldots,L^{(1)}.
\end{equation}
Subsequently for frames-$2,3,4,\ldots$, tracks are updated using Eq.~(\ref{eqn:track_history}) and Eq.~(\ref{eqn:track_update}). Further, a user can, at any intermediate frame, mark a segment to initialize a new track.
Any track-$i$ ceases to exist when it does not find any segment with high feature similarity using Eq.~(\ref{eqn:track_update}), as: 
\begin{equation}
\label{eqn:track_cease}
\Phi({H_i^{(k-1)}, \mathrm{s}_j}) < \varepsilon,\, \forall \mathrm{s}_j \in \mathrm{R}_b^{(k)},
\end{equation}
where $\varepsilon$ is a very small threshold. This method is illustrated in Fig.~\ref{fig:tracking}.
After processing all frames, segments belonging to the detected tracks jointly constitute the \emph{Body} region of the mummy $\mathrm{R}_b$, which is the most challenging region to separate.
The visualization of the segmented body is shown in Fig.~\ref{fig:visualize_body}.

\section{Results}
\subsection{Parameters}
In our experiments, we consider $m = 10$ geodesic distances in Eq.~(\ref{eqn:avg_dist_top}) for computing average geodesic distance. In the experiments for the GrabCut stage (Sec.~\ref{ssec:stage1}), we choose overlapping volumes of  $n_G = 10$ successive  frames with stride of 1, to avoid any loss of information. For higher resolution data, a bigger value of $n_G$ can be chosen, since statistics change slowly across successive frames compared to low resolution data.
During the tracking stage (Sec.~\ref{ssec:stage2}), we set the parameter for storing track history $M = 4$. Like $n_G$, a bigger value can be chosen in case of higher resolution data.
\begin{figure}[t]
    \centering
    \subfloat[]{
    \includegraphics[width = 0.5\linewidth, clip, trim={2mm 0 0 6mm}]{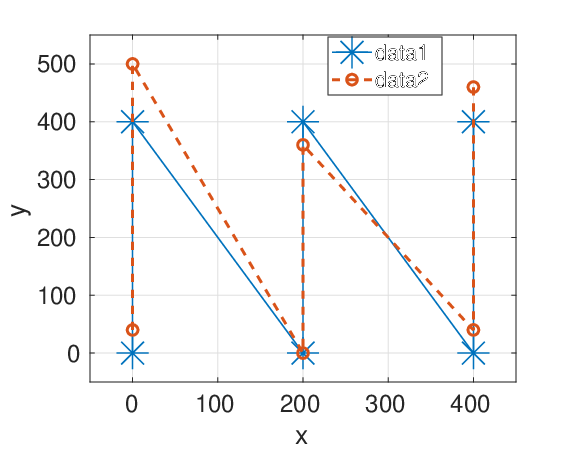}
    \label{subfig:warp1}}
    \subfloat[]{
    \includegraphics[width = 0.5\linewidth, clip, trim={0 0 0 6mm}]{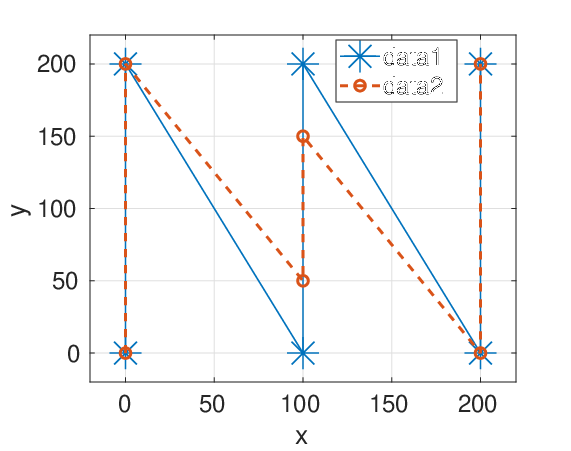}
    \label{subfig:warp2}} \\
    \subfloat[]{
    \includegraphics[width = 0.25\linewidth]{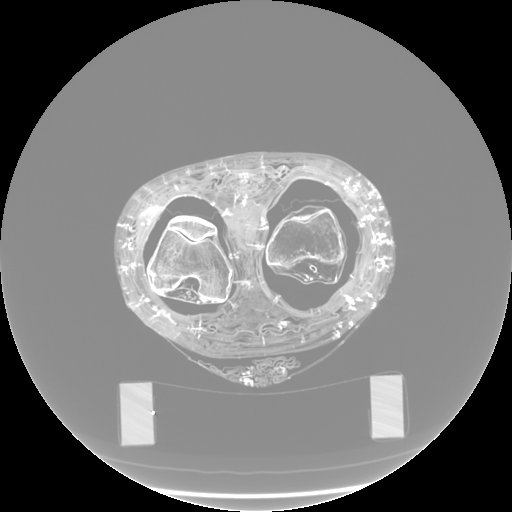}
    \label{subfig:input_img}} ~
    \subfloat[]{
    \includegraphics[width = 0.25\linewidth]{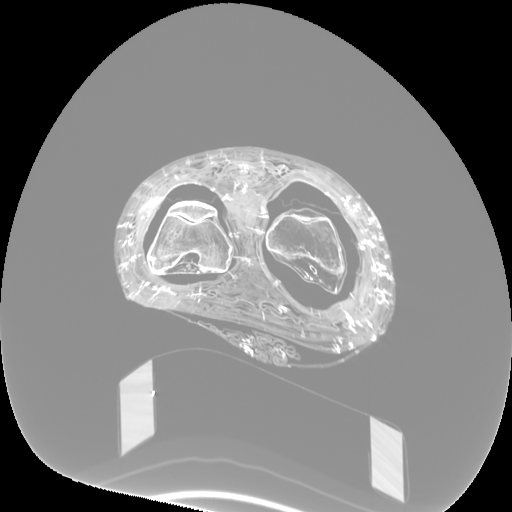}
    \label{subfig:warp1_img}} ~
    \subfloat[]{
    \includegraphics[width = 0.25\linewidth]{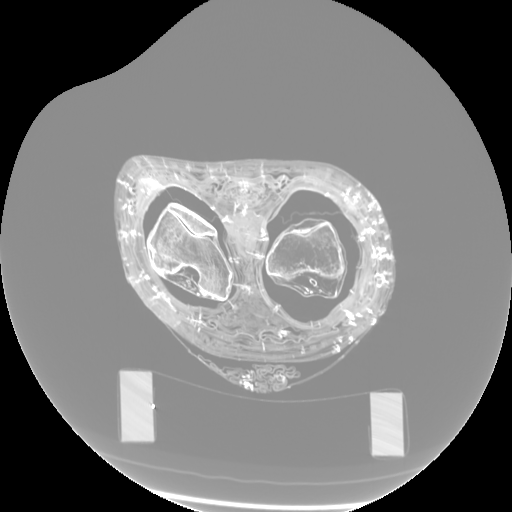}
    \label{subfig:warp2_img}}
    \caption{Illustration of transformation of a CT data to generate several new data. (a) Two sets of points for which a warping function is computed using thin-plate spline mapping. (b) A different pair of point sets. (c) An axial frame of the CT data. (d,e) The frame in (c) is transformed using the warping functions computed using point sets in (a) and (b), respectively.}
    \label{fig:warping_img}
\end{figure}
\begin{figure}[b]
    \centering
    \includegraphics[width = 0.64\linewidth]{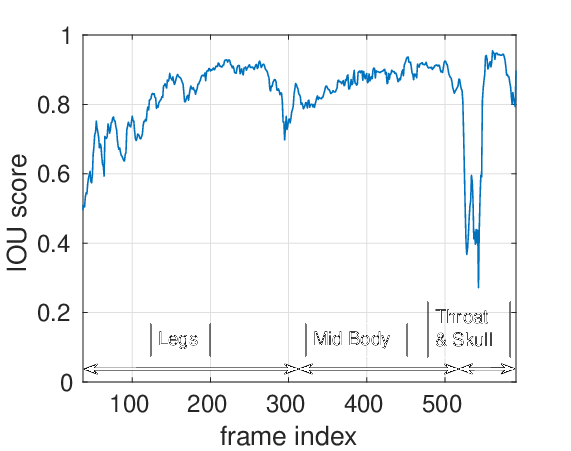}
    \caption{Intersection over union (IOU) score computed for every frame. The frames are of feet to the skull of the mummy as we move from lower frame index to higher frame index.}
    \label{fig:iou}
\end{figure}
%

\subsection{Dataset}
We have tested our approach on CT scan of a human mummy. In the absence of a large number of subjects for validating the proposed method, we generated additional CT scans by transforming the original scan. For this purpose, we used thin-plate spline mapping~\cite{bookstein:principal}, where we learn a warping function between two sets of points in the XY-plane and apply it on the coordinates of the input data to transform it. 

We first select a set points $X_1 = (x_1,y_1)$, $X_2 = (x_2,y_2)$, $\ldots$, $X_n = (x_n,y_n)$ in the XY-plane. Then they are perturbed slightly to obtain another set of points $X_1\p = (x_1\p,y_1\p)$, $X_2\p = (x_2\p,y_2\p)$, $\ldots$, $X_n\p = (x_n\p,y_n\p)$ (see Fig.~\ref{subfig:warp1}). Then using thin plate spline, we learn the warping function $f(x,y)$ that maps $X_i$'s to $X_i\p$'s with the minimum bending energy~\cite{PMT}. Next, for each frame in the CT data, $f(x,y)$ is used to map each pixel to the new coordinates and a warped frame is obtained through interpolation of pixels values. 
%
%
This method is particularly flexible, since the initial set of points $X_i$'s and their perturbations can be chosen arbitrarily. Hence in theory, we can obtain an infinite number of warping functions. 
Fig.~\ref{fig:warping_img} shows the outputs generated by warping an input frame  with two different warping functions.

In the above strategy, every frame in the CT data is transformed using the same warping function. We further investigate the effect of different transformation for every frame. Let $N$ be the number of frames in the CT data and denote $X_i$'s as $X_i^{(0)}$ to $X_i\p$'s as $X_i^{(N)}$. For each frame-$k = 1, 2, \ldots, N$, we compute 
\begin{equation}
\label{eqn:warp_set1}
    X_i^{(k)} = X_i^{(0)} + k (X_i^{(N)} - X_i^{(0)}) / N 
\end{equation}
and obtain the corresponding warping function $f_k(x,y)$ using point sets $X_i^{(k-1)}$ and $X_i^{(k)}$. Let us call this set of warping functions as Set 1: $\{ f_k(x,y) \}$. To avoid biasing different section of the mummy to different degree of transformations, we also obtain three sets by permuting $\{ f_k \}$ as follows.
\begin{equation}
\label{eqn:warp_set2}
    \text{Set 2: } \{ f_N, f_{N-1}, \ldots, f_1 \}
\end{equation}
\begin{equation}
\label{eqn:warp_set3}
    \text{Set 3: } \{ f_1, f_2, \ldots, f_{N/2}, f_{N/2}, \ldots,  f_2, f_1 \}
\end{equation}
\begin{equation}
\label{eqn:warp_set4}
    \text{Set 4: } \{ f_{N/2}, f_{N/2-1}, \ldots, f_1, f_1, \ldots, f_{N/2-1}, f_{N/2} \}
\end{equation}
\begin{table}[b]
    \caption{IOU Score on Different Data and Ablation Study.}
    \label{tab:iou}
    \setlength{\tabcolsep}{1.4mm}
    \centering
    \begin{tabular}{||c||c|c|c|c|c|c|c||}
     \hline
     & Original & Warp 1 & Warp 2 & Set 1 & Set 2 & Set 3 & Set 4 \\
    \hline\hline
    Legs    & 0.79 & 0.77 & 0.77 & 0.77 & 0.77 & 0.77 & 0.77 \\
    \hline
    Mid-body& 0.87 & 0.88 & 0.88 & 0.88 & 0.88 & 0.88 & 0.89 \\
    \hline
    Head    & 0.74 & 0.79 & 0.81 & 0.80 & 0.79 & 0.79 & 0.79 \\
    \hline\hline
    \textbf{Average} & 0.81 & 0.81 & 0.81 & 0.82 & 0.81 & 0.82 & 0.82 \\
    \hline
    Average-& & & & & & & \\
    w/o tracking & 0.79 & 0.80 & 0.81 & 0.80 & 0.80 & 0.80 & 0.80\\
    \hline
    \end{tabular}
\end{table}
%

\subsection{Quantitative result}
\label{ssec:quanti}
To quantitatively measure the performance of the proposed method, we use intersection over union (IOU) between the predicted result and the manually annotated ground-truth available for a single mummy CT scan. IOU has been computed both per frame and on the full mummy volume.

The plot of the IOU score computed for each mummy frame is shown in Fig.~\ref{fig:iou}. Poor accuracy for the initial few frames is due to similar visual appearances of the \emph{Bandages} and the \emph{Body} region voxels which cannot be improved using the tracking method. The drop in accuracy near frame-540 is due to the presence of the metallic necklace which caused severe artifacts during data acquisition, making it impossible to recover the correct voxel radiodensity in certain regions.

We analyze the average IOU score over the full mummy volume and on different parts of the body (legs, mid-body and head). Table~\ref{tab:iou} contains the results for the input data and six transformed data obtained using:
\begin{enumerate}[label=\roman*)]
\item the warping function of Fig.~\ref{subfig:warp1};
\item the warping function of Fig.~\ref{subfig:warp2};
\item 4 warping function sets (Eq.~(\ref{eqn:warp_set1})-(\ref{eqn:warp_set4})) obtained from (i).
\end{enumerate}
\begin{figure}[b]
    \centering
    \includegraphics[width = 0.52\linewidth]{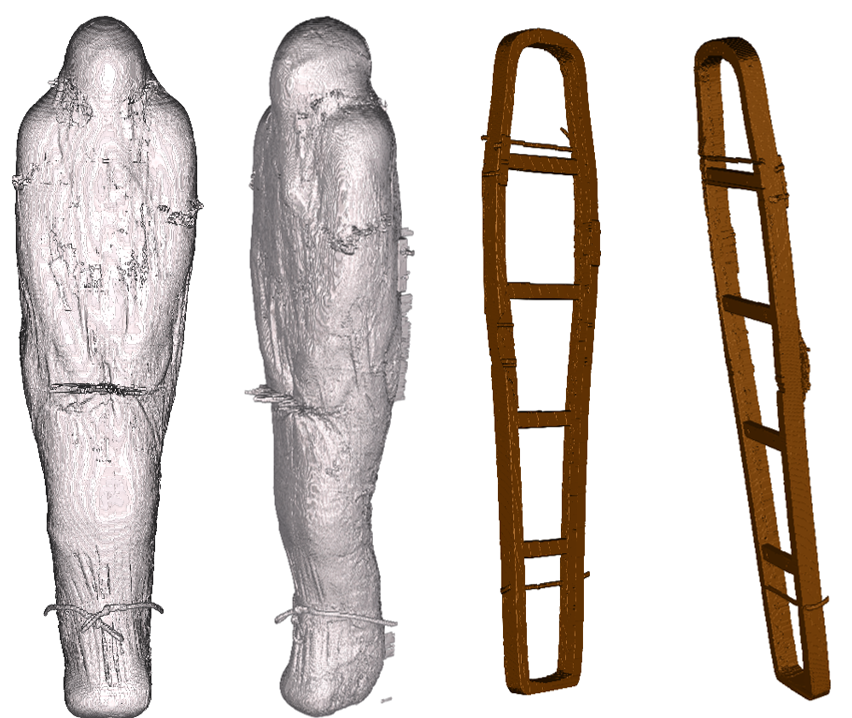}
    \caption{Visualization from different viewpoints of the entire wrapped mummy and the support as detected using the proposed method.}
    \label{fig:visualize_wrap}
\end{figure}

Ground-truth mask of the input data is also transformed using the respective warping functions for their IOU score computation. 
An ablation study of the proposed method is also given in last row of Table~\ref{tab:iou}, which reports the performance of the proposed method without using the tracking algorithm. We observe that the tracking algorithm is indeed beneficial.

Here, we make a comment about the challenges faced. It is a common occurrence that some part of the \emph{Body} and its adjacent wrapping \emph{Bandage} have similar radiodensity. This makes it very difficult to separate them accurately. After transforming the data using warping functions, it is possible that the amount of adjacency between \emph{Bandage} and \emph{Body} with similar radiodensity changes. Accordingly, the performance changes. Further, reflections from metals introduce artifacts in the data by changing (both increasing and decreasing) certain voxel values (Fig.~\ref{fig:artifact}), thus providing ambiguous information about density of different tissues and affecting the performance.
%
%

\textbf{Comparative study:} We compare the performance of the proposed method with existing semi-supervised segmentation methods by evaluating them on the mummy scan (Table~\ref{tab:compare}).

We consider a Graph Cut based clustering method~\cite{borovec2017supervised} (referred as \textit{GC}) that had been applied to medical image segmentation. It learns a Gaussian mixture model for a predefined number of clusters ($n_C$) and assigns each voxel to one of them. The cluster having the highest overlap with the ground-truth mask is chosen for computing IOU. Further, to ensure fairness, we provide supervision to the algorithm by additionally marking the exterior region (this supervision is also provided to every method we compare with). In Table~\ref{tab:iou_gc}, we report the IOU scores obtained by varying $n_C$ between 3-10 and observe that results do not vary significantly.
\begin{table}[t]
    \caption{IOU Score of Graph Cut Based Segmentation Method (GC) \cite{borovec2017supervised} Considering $n_C$ Classes.}
    \label{tab:iou_gc}
    \centering
    \begin{tabular}{||c||c|c|c|c|c|c|c|c||}
         \hline
         $n_C$ &\textit{3} &\textit{4} &\textit{5} &\textit{6} &\textit{7} &\textit{8} &\textit{9} &\textit{10} \\ 
         \hline\hline
         GC &0.41 &0.42 &0.41 &0.42 &0.42 &0.41 &0.40 &0.39 \\
         \hline
    \end{tabular}
\end{table}

Next, we apply the standard GrabCut method~\cite{rother2004grabcut} (referred as \textit{GB}) on the input data by drawing bounding boxes around the \emph{Body} region. The achieved performance (Table~\ref{tab:compare}) is much inferior to that of our proposed method. This is because our efficient geodesic segmentation output acts as a much better input label for GrabCut. 
We also investigated how salient the mummy \emph{Body} region is. For this, we have computed saliency map of each frame using the method in~\cite{zhu2014saliency}  (referred as \textit{SD}). The performance obtained is poor because of the very small contrast between tissues in the \emph{Body} and \emph{Bandages}. 
Finally, we evaluated a video object segmentation method~\cite{griffin2019tukey} (referred as \textit{TIS}) to investigate its applicability on the mummy 2D image sequence. This method uses optical flow and visual saliency of frames, but it is unable to track the mummy's \emph{Body}.
\begin{table}
    \caption{Comparison of IOU Score using the Proposed Method (\textit{Our}), GrabCut Segmentation (\textit{GB}) \cite{rother2004grabcut}, Saliency Detection (\textit{SD}) \cite{zhu2014saliency}, Graph Cut Segmentation (\textit{GC}) \cite{borovec2017supervised} and Video Object Segmentation \textit{TIS} \cite{griffin2019tukey}.}
    \label{tab:compare}
    \centering
    \begin{tabular}{||c|c|c|c|c|c||}
        \hline
         & \textit{Our} & \textit{GB} & \textit{SD} & \textit{GC} & \textit{TIS}  \\
         \hline\hline
         IOU Score & 0.81 & 0.50 & 0.46 & 0.42 & 0.24 \\
         \hline
    \end{tabular}
\end{table}
\begin{figure}
    \centering
    \includegraphics[width = 0.85\linewidth]{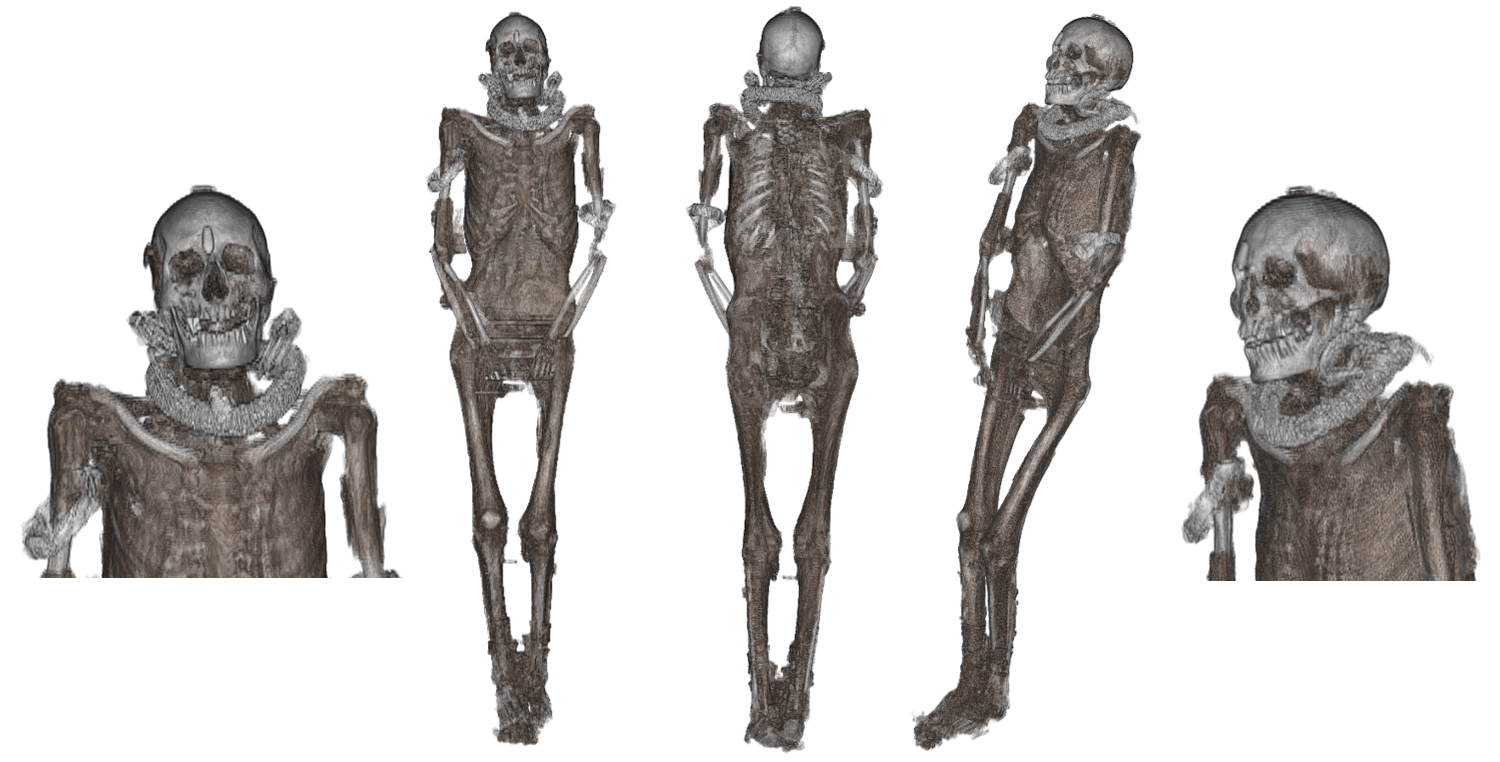}
    \caption{Visualization from different viewpoints of the body region detected using our method and zoomed visualizations of skull area.}
    \label{fig:visualize_body}
\end{figure}
\begin{figure}
    \centering
    \includegraphics[width = 0.45\linewidth]{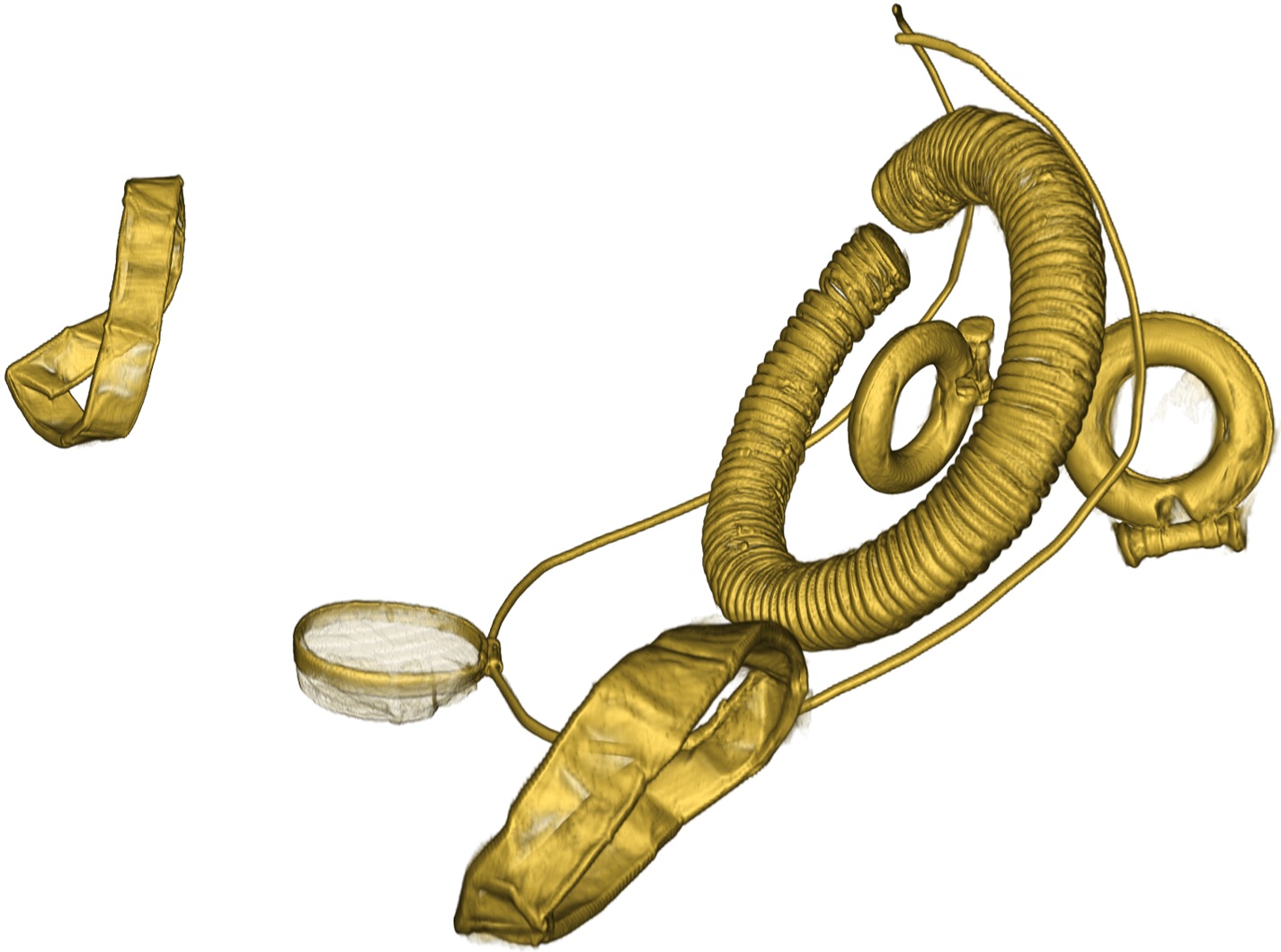}
    \caption{Visualization of metals (jewelry) inside the mummy.}
    \label{fig:visualize_gold}
\end{figure}
\begin{figure}[t]
    \centering
    \includegraphics[width = 0.7\linewidth]{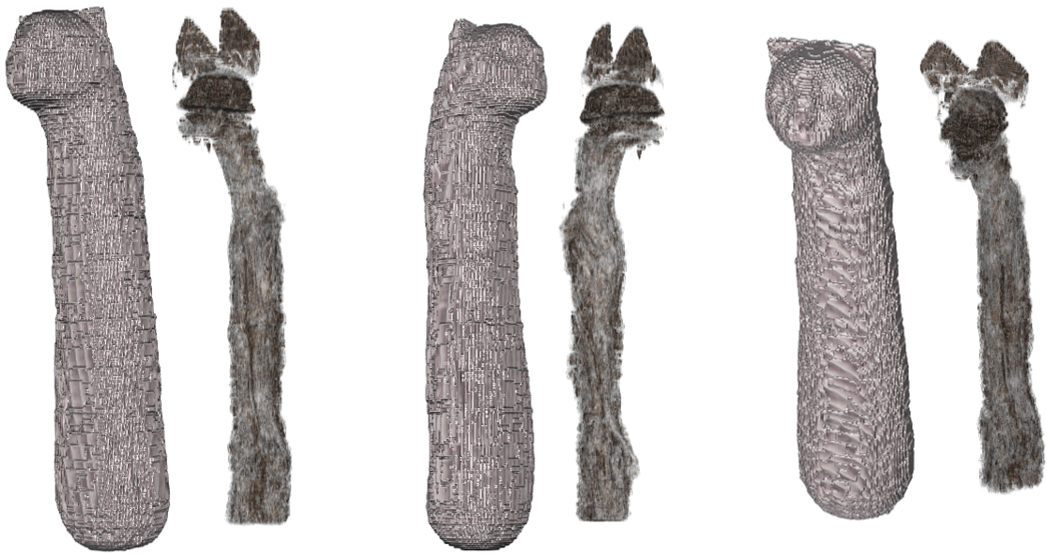}
    \caption{Visualization from different viewpoints of the wrapped cat mummy and the corresponding segmented body.}
    \label{fig:visualize_catwrap}
\end{figure}
%

\subsection{Qualitative result}
\label{ssec:quali}
In this section, we visualize the produced segments.
The pre-processing stage separates the voxels belonging to the wrapped mummy from the the external space and the exterior objects (mummy support). Fig.~\ref{fig:visualize_wrap} shows the result from different viewpoints. 
Near the fingers, arms and the neck of the mummy, some scattering can be observed. These are the artifacts caused by the presence of metals in the form of jewelries, as discussed in Sec.~\ref{ssec:data} and highlighted in Fig.~\ref{fig:artifact}.

Fig.~\ref{fig:visualize_body} shows the unwrapped mummy's body, including the skull obtained after the final stage of the segmentation pipeline.

In Fig.~\ref{fig:visualize_gold}, we show the jewelries present in the mummy, segmented in the pre-processing stage. 

Though the pipeline is explained using a human mummy, it can also be applied to other mummies. In Fig.~\ref{fig:visualize_catwrap}, we present the results obtained by applying our method on a cat mummy.

\section{Conclusion}
In this paper, we proposed an algorithm for solving the problem of segmenting mummy CT scan with minimal supervision. Leveraging the unique structure of the data, we applied geodesic distance measure to obtain an initial segmentation, which we further refined using GrabCut and tracking. In order to validate our method on multiple CT scans, we proposed to use thin-plate spline mapping to transform the original data and artificially generate additional samples. The proposed pipeline yielded very good results using minimal user interactions compared to existing semi-supervised methods. The presence of artifacts in the data affects the segmentation results. Overcoming this difficulty is  an ongoing research.

\section*{Acknowledgements}
This research has been developed in the context of the collaboration between Fondazione Istituto Italiano di Tecnologia and Fondazione Museo delle Antichità Egizie di Torino. We thank our colleagues Enrico Ferraris and Marco Rossani from Museo Egizio di Torino, who provided their insights and expertise in cultural heritage research.

\bibliographystyle{IEEEtran}
\bibliography{IEEEabrv,ref}
%
\end{document}